\documentclass[runningheads]{llncs}
\usepackage{graphicx}
\usepackage{caption}
\usepackage{makecell}
\usepackage{booktabs}
\usepackage{multirow}
\usepackage{tabularx}
\usepackage{hyperref} 

\usepackage{glossaries}
\setacronymstyle{long-short}
\newacronym{ML}{ML}{machine learning}
\newacronym{XAI}{xAI}{explainable artificial intelligence}
\newacronym[longplural={counterfactual explanations}]{CFE}{CFE}{counterfactual explanation}
\newacronym{SCS}{SCS}{System Causability Scale}

\usepackage{xspace}
\newcommand*{\eg}{e.g.\@\xspace}
\newcommand*{\ie}{i.e.\@\xspace}
\usepackage{amssymb}
\usepackage{braket}
\newcommand{\x}{\ensuremath{\vec{x}}}
\newcommand{\xcf}{\ensuremath{{\vec{x}_{\text{cf}}}}}
\newcommand{\RN}{\mathbb{R}}
\newcommand{\dimsym}{d}
\newcommand{\classifier}{\ensuremath{h}}
\newcommand{\setY}{\ensuremath{\set{Y}}}
\DeclareMathOperator*{\loss}{{\ell}}
\newcommand{\ycf}{\ensuremath{y'}}
\DeclareMathOperator*{\regularization}{\ensuremath{{\theta}}}
\newcommand{\refeq}[1]{Eq.~\eqref{#1}}

\usepackage{comment}

\begin{document}
\title{For Better or Worse: The Impact of Counterfactual Explanations' Directionality on User Behavior in xAI\thanks{This research was supported by research training group Dataninja (Trustworthy AI for Seamless Problem Solving: Next Generation Intelligence Joins Robust Data Analysis) funded by the German federal state of North Rhine-Westphalia, and by the European Research Council (ERC) under the ERC Synergy Grant Water-Futures (Grant agreement No. 951424).}}
\titlerunning{The impact of CFE directionality on user behavior in xAI}
%
\author{Ulrike Kuhl\inst{1,2}\orcidID{0000-0002-9405-918X} \and
    André Artelt\inst{2}\orcidID{0000-0002-2426-3126} \and
    Barbara Hammer\inst{2}\orcidID{0000-0002-0935-5591}}
\authorrunning{U. Kuhl et al.}
%

\institute{Research Institute for Cognition and Robotics (CoR-Lab), Bielefeld University, Bielefeld, Germany\\
\url{https://www.uni-bielefeld.de/zwe/cor-lab/}\and
Machine Learning Group, Faculty of Technology, Bielefeld University, Bielefeld, Germany\\
\url{https://hammer-lab.techfak.uni-bielefeld.de/doku.php}\\
\email{\{ukuhl,aartelt,bhammer\}@techfak.uni-bielefeld.de}}
\maketitle              
\vspace{-1cm}
\begin{abstract}

\Glspl{CFE} are a popular approach in \gls{XAI}, highlighting changes to input data necessary for altering a model's output.
A \gls{CFE} can either describe a scenario that is better than the factual state (\textit{upward} \gls{CFE}), or a scenario that is worse than the factual state (\textit{downward} \gls{CFE}).
However, potential benefits and drawbacks of the directionality of \glspl{CFE} for user behavior in \gls{XAI} remain unclear.
The current user study (N=161) compares the impact of \gls{CFE} directionality on behavior and experience of participants tasked to extract new knowledge from an automated system based on model predictions and \glspl{CFE}.
Results suggest that \textit{upward} \glspl{CFE} provide a significant performance advantage over other forms of counterfactual feedback. Moreover, the study highlights potential benefits of \textit{mixed} \glspl{CFE} improving user performance compared to \textit{downward} \glspl{CFE} or no explanations. In line with the performance results, users' explicit knowledge of the system is statistically higher after receiving \textit{upward} \glspl{CFE} compared to \textit{downward} comparisons.
These findings imply that the alignment between explanation and task at hand, the so-called regulatory fit, may play a crucial role in determining the effectiveness of model explanations, informing future research directions in \gls{XAI}.
To ensure reproducible research, the entire code, underlying models and user data of this study is openly available: \url{https://github.com/ukuhl/DirectionalAlienZoo}


\begin{minipage}[]{0.2\textwidth}
\includegraphics[scale=0.7]{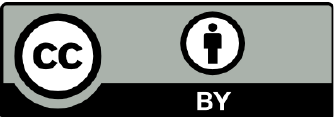}
\end{minipage} \hfill
\begin{minipage}{0.6\textwidth}
This work is licensed under a Creative Commons  \href{https://creativecommons.org/licenses/by/4.0/}{``Attribution 4.0 International''} license.
\end{minipage}

\end{abstract}
%
%
\glsresetall
%
%


\section{Introduction}

The question of how to provide users with understandable, usable, and trustworthy explanations for \gls{ML} decisions is at the heart of \gls{XAI}.
A popular variant within the community are \glspl{CFE}, drawing out ``what-if'' scenarios that highlight necessary perturbations of the input data to change a model's output~\cite{wachter2017counterfactual}.
Recent years have brought a notable uptick of studies investigating various aspects of \glspl{CFE} for \gls{ML}.
Prior work focuses, inter alia, on their robustness~\cite{artelt2021evaluating}, impact on user trust and satisfaction~\cite{warren2022features,warren2023categorical}, and usability as a function of algorithmic properties~\cite{kuhl2022keep}(see~\cite{verma2020counterfactual} for an extensive review of the research landscape).

One key defining characteristic of counterfactual statements is their directionality: \textit{upward} counterfactuals describe scenarios that are superior to the factual state (\ie, how it would have been better), while \textit{downward} counterfactuals refer to more negative alternatives to the factual state (\ie, how it would have been worse)~\cite{markman1993mental}.

There is general agreement among cognitive and social psychologists that \textit{upward} counterfactuals serve a preparatory role, increasing motivation and guiding future action~\cite{byrne2016counterfactual,wong2007narrating}.
The role of \textit{downward} counterfactuals, however, seems to be more complex.
A common argument points towards a predominantly affective role, inducing a sense of relief about the factual state by emphasizing how a scenario could have been worse~\cite{roese1995functions}.
However, alternative empirical evidence suggests that \textit{downward} counterfactuals may act as a wake-up call by drawing attention towards the possibility of worse outcomes, thus increasing motivation to take action~\cite{mcmullen2000downward}.

In \gls{XAI}, the impact of \gls{CFE} directionality remains even more ambiguous, given that counterfactuals used to explain a model are not spontaneously generated by humans, but automatically computed as actionable feedback deepening users' understanding.
\gls{CFE} user studies commonly investigate \glspl{CFE} that flip a binary outcome class~\cite{dai2022counterfactual,celar2023people,warren2022features,warren2023categorical,ramon2021understanding}. 
While these outcomes may have qualitative implications within their respective task domains (\eg, being under vs. over the legal blood alcohol limit to drive~\cite{warren2022features,celar2023people,warren2023categorical}, chemicals being safe vs. unsafe~\cite{celar2023people}, grass growth levels on a farm being high vs. low~\cite{dai2022counterfactual}), the directionality of provided explanations is often outside the respective research focus. 
Thus, this aspect has not yet been extensively studied in \gls{XAI}, and preliminary data available presents inconsistencies. 

For instance,~\cite{ramon2021understanding} report \textit{downward} \glspl{CFE} for positive decisions to be less popular compared to importance rankings, and find no differences between rankings and \textit{upward} \glspl{CFE} in terms of user preference. 
In contrast,~\cite{celar2023people} suggest a behavioral impact of explanations that establish a \textit{downward} comparison to the factual state on personal decision-making. 
Following this reasoning, \textit{downward} \glspl{CFE} may potentially serve as better actionable feedback.

Given these sparse and inconsistent accounts, it is unclear whether one type of \glspl{CFE} 
is more effective than the other in improving user performance in tasks that require model interpretation.
Therefore, the current study systematically compares the impact of \gls{CFE} directionality for \gls{ML} predictions on user behavior.
Specifically, we perform a user study that requires participants to extract new knowledge from an automated system given model predictions and corresponding \glspl{CFE}.
On top of groups exclusively receiving \textit{upward} and \textit{downward} \glspl{CFE}, we provide a third group of users with both types in a \textit{mixed} condition. 
We find it conceivable that collective information on better and worse outcomes may grant a more complete understanding of the causal relationships between actions and outcomes, effectively informing future decision-making.
We investigate how \glspl{CFE} of either type impact users' objective performance, explicit knowledge of the system, and subjective experience, compared to each other and a no-explanation \textit{control} condition.

\section{Related Work}

In contrast to using inherently interpretable models such as rule sets or decision trees, establishing explainability for opaque models like support vector machines or neural networks is a challenging endeavor.
Proposed approaches include feature importance methods providing insights into the relevance and influence of input features on model predictions~\cite{lundberg2017unified,rozemberczki2022shapley}, rule extraction techniques distilling interpretable decision rules from complex models~\cite{qiao2021learning}, and prototype-based explanations leveraging representative instances to explain model behavior~\cite{shin2022prototype}.

In this broader \gls{XAI} landscape, \glspl{CFE} take a prominent role given an emerging user-centric focus on explainability~\cite{miller_explanation_2019}.
\glspl{CFE} facilitate human comprehension by explicitly revealing the necessary changes in input data to influence the model's output~\cite{wachter2017counterfactual}. 
In this way, \glspl{CFE} provide explanations for instances where the model's predictions deviate from the desired outcomes, allowing users to understand the factors that contribute to the model's decision-making process.
Their particular appeal lies in their intrinsically contrastive format, bearing a strong resemblance to human cognitive reasoning.
Indeed, individuals routinely engage in counterfactual thinking~\cite{roese_counterfactual_1997,goldinger_blaming_2003}.
During this process, one not only retains the representation of actual facts, but also simulates an alternative scenario of how the reality might have differed~\cite{byrne2016counterfactual}.
This distinct characteristic positions \glspl{CFE} as a valuable addition to the \gls{XAI} toolkit, promising to provide users with actionable insights for decision-making and understanding model behavior for a given decision.

How humans reason from counterfactuals has been a prominent research topic in cognitive psychology studies~\cite{sanna_antecedents_1996,hilton_knowledge-based_1986,lombrozo_explanation_2012}, producing relevant implications for the use of \glspl{CFE} in \gls{XAI} applications~\cite{byrne_counterfactuals_2019}.
Human-generated counterfactuals typically change only a limited set of features, preferably undoing recent and controllable events to create hypothetical scenarios that are strongly aligned with the individual's personal world knowledge and beliefs~\cite{dyczewski2012general,byrne_counterfactuals_2019}.
The current literature encompasses various computational approaches for generating \glspl{CFE}~\cite{stepin2021survey,artelt2019computation}, reflecting the continuing development within the field of counterfactual explanation generation.
To yield explanations that closely resemble human counterfactual thought~\cite{miller_explanation_2019}, generation approaches have placed emphasis on producing \glspl{CFE} that are sparse~\cite{mothilal2020explaining}, stay close to the original input (with variations in terms of the distance measures,~\cite{keane2020good,karimi2020model}), focus on controllable (and thus actionable) features~\cite{ustun2019actionable,karimi2020model}, and may even diversify the generated solutions to meet end-user's needs~\cite{mothilal2020explaining}.

Still, gaps in understanding in how far certain aspects of \glspl{CFE} may facilitate or hinder a user's understanding when they are used in \gls{XAI} remain.
Just recently,~\cite{warren2023categorical} demonstrated that human users more readily understand explanations relying on categorical features in contrast to continuous ones, a distinction not typically taken into account by \gls{CFE} generation approaches.
In a similar vein, the current work investigates the potential impact of \gls{CFE} directionality on user behavior, a fundamental property commonly not addressed in \gls{XAI}.
\textit{Upward} counterfactuals (\ie, how it would have been better) are typically generated following negative events~\cite{markman1993mental}. 
In this way, they may provide a clear roadmap for future improvement and action~\cite{byrne_counterfactuals_2019}.
Indeed, imagining ``better-worlds'' broadly leads to performance improvements in various tasks and settings as a driving force for learning from past mistakes~\cite{roese1994functional,wong2007narrating,epstude2008functional,myers2014role}.
When individuals engage in \textit{upward} counterfactual thinking, their motivational orientation towards improvement aligns with the counterfactual focus on a hypothetical ``better world'', thus inducing regulatory fit~\cite{higgins2000making}.
The positive affect associated with regulatory fit may enhance motivation, persistence, and goal-directed behavior, leading to an increased likelihood of taking action to bridge the gap between the current and desired states~\cite{motyka2014regulatory}.

\textit{Downward} counterfactuals, in contrast, refer to imagining more negative alternatives to the factual state (\ie, how it would have been worse)~\cite{markman1993mental}.
This downward comparison may have different functional implications, and research indeed reveals a complex pattern.
On the one hand, \textit{downward} counterfactual thinking is frequently associated with affective regulation, eliciting relief~\cite{roese1995functions} and reducing regret~\cite{parikh2022efficacy}.
Through this positive affective role of inducing a feeling of ``I’m better off than I could have been'', it seems to serve a self-enhacement function leading to more favorable self-perception~\cite{white2005looking}.
In this way, \textit{downward} counterfactual thinking may lead to a sense of complacency, reducing the motivation to act~\cite{mcmullen2000downward}.
On the other hand, putting one's attentional focus on an objectively worse counterfactual possibility may induce negative affect, which may in turn serve as a motivator by signaling that the present condition is inadequate and requires action~\cite{mcmullen2000downward}.
Thus, by focusing on mistakes and missed opportunities, \textit{downward} counterfactuals may potentially highlight areas for improvement.
Despite these indications for fundamental differences in how humans reason with \textit{upward} and \textit{downward} counterfactuals, this crucial aspect of \glspl{CFE}' effectiveness and usability received little attention in \gls{XAI} research so far.
An extensive literature review revealed only two previous papers partially addressing this issue. 

First, \cite{ramon2021understanding} conducted a study examining the effectiveness of feature rankings and \glspl{CFE} in two everyday contexts: online advertising and loan applications. Specifically, their second experiment focuses on the directionality of explanations, with a particular emphasis on providing \textit{upward} CFEs following negative outcomes and \textit{downward} CFEs following positive outcomes.
Participants made trade-off decisions between the two explanation modes, thus indicating their preferences for either feature rankings or \glspl{CFE}. The results present a notable contrast to the prevailing preference for \glspl{CFE} within the \gls{XAI} community. Surprisingly, users show a higher preference for feature rankings over \textit{downward} \glspl{CFE} when faced with positive outcomes. This suggests that users found feature rankings more favorable in such scenarios.
In the case of negative decisions, users exhibit no significant difference in terms of preference between the explanation formats, selecting \textit{upward} \glspl{CFE} as frequently as feature rankings.
It is important to note, however, that such an assessment of user preference does not specifically allow drawing conclusions about relative usability differences of the explanation formats. Usability and user preference are two distinct aspects when evaluating the effectiveness of a system; aspects that -- while being associated to some extent -- often do not align~\cite{nielsen1994measuring}. Users may exhibit a subjective preference for systems or explanations, irrespective of the measurable impact on performance~\cite{van2021evaluating,lage2019human}.

More recently,~\cite{celar2023people} exposed participants to a model's input, its decision, and either counterfactual or causal explanations, framed as a software application built to aid decision-making.
Depending on the experimental group, the domain presented to a participant encompassed either a familiar scenario (\ie, blood alcohol level and driving limit), or an unfamiliar one (\ie, chemical safety).
After rating the perceived helpfulness of the explanation presented, participants reached personal decisions whether they would be prepared to drive / handle an unknown chemical for a series of cases where they only saw the model input (Experiment 2 of~\cite{celar2023people}). Intriguingly, the personal decisions aligned better with model predictions when the preceding explanations would specifically establish a downward comparison.
While this may shed a favorable light on \textit{downward} \glspl{CFE} for guiding personal action, it is unclear to which extent the familiarization phase framed as judging the helpfulness of a software application carried over to the subsequent decision-making phase. 
Furthermore, the reported beneficial effect of explanations that establish a downward comparison presents an incidental finding, as the actual focus of the study was on effects related to domain familiarity.  

In light of these inconclusive preliminary findings, we aim to take a first step towards a systematic investigation of their directionality impacts \glspl{CFE}' usability as actionable feedback in \gls{XAI}.
Specifically, we ask whether novice users tasked to gain new information from an unknown system in an abstract domain~\cite{kuhl2023let} benefit more from receiving \textit{upward}, \textit{downward}, or \textit{mixed} \gls{CFE} feedback.
By examining the effects of directionality, we aim to shed light on a nuance of \glspl{CFE} that has yet to be explored, contributing to a more comprehensive understanding of their effectiveness and applicability in \gls{XAI}.

\section{Methods}

To assess the impact of directionality of \glspl{CFE} in \gls{XAI} on user behavior, we employ the game-inspired Alien Zoo framework~\cite{kuhl2023let}.
Consequently, our study assesses the efficacy of \textit{upward}, \textit{downward}, and \textit{mixed} CFEs in acquiring new knowledge from an automated system in a low-knowledge domain, specifically targeting novice users.
The Ethics Committee of Bielefeld University, Germany, approved this study.

\subsection{Participants}

We determined the required sample size for the current study by running an \textit{a-priori} power analysis, using openly-available empirical data from an earlier empirical study based on the same experimental paradigm~\cite{kuhl2023let}.
These exemplary data provided us with realistic estimates for fixed and random effects to be expected in the current study.
The power analysis (R package mixedpower v.0.1.0~\cite{kumle2021estimating}) indicated that 40 participants per group were required to achieve a power of \textgreater85\% (medium effect size with alpha\textless.05).

161 Participants were recruited in April 2023 using \textit{Prolific Academic}\footnote{\url{https://www.prolific.co/}}, and assigned to one of four between-participant conditions in a fixed order: \textit{upward} \glspl{CFE} (n=40), \textit{downward} \glspl{CFE} (n=40), \textit{mixed \glspl{CFE}} (\ie, receiving one \textit{downward} and one \textit{upward} \glspl{CFE} in each feedback round, n=41), and a no-explanation \textit{control} group (n=40). 
We restricted access to the study to native English speakers from the United States, Australia, Canada, New Zealand, Ireland, and the United Kingdom, who did not previously participate in studies with the given experimental framework.
Before participating, users provided informed consent through electronic click wrap agreement.

All participants received a base pay of GBP\textsterling 4 for participation.
The three top performers in each condition received a bonus payment of GBP\textsterling 1.
Together with the experimental instructions, participants were informed about a potential monetary bonus to increase compliance with the task~\cite{bansal_updates_2019}.

To ensure sufficient data quality, we applied several exclusion criteria prior to analysis.
Specifically, a participant's responses were removed when they failed more than one attention check (n=3). No participant displayed monotonous response patterns despite poor performance during the game, indicating high effort. 
Data from 158 participants contributed to the final game analysis (Table \ref{tab:Participants}).

Note that for a number of users, logging of survey responses failed due to technical difficulties (n=8). Further, we excluded users from the subsequent survey analysis if they answered with positive or negative valence only, indicating low-effort entries (n=4). Consequently, the survey analysis was based on a subset of 146 users from the game phase.

\begin{table}
  \caption{Demographic information of participants.}
  \label{tab:Participants}
\begin{tabularx}{\textwidth}{p{2cm}XXXlXX} 
\toprule
    & \multicolumn{6}{l}{Before quality assurance measures (\textit{N} = 161)}\\
\cline{2-7}
    & \textit{control} & \textit{upward} & \textit{downward} & \textit{mixed} & \textit{H} value$^a$ & \textit{p} value\\ 
\hline
\textit{N}   &  40 & 40 & 40 & 41 & .. & ..\\
Gender$^{b,c}$ & 20f/20m & 20f/20m & 20f/20m & 20f/19m/1nb & 0.018 & .999 \\
Age (\textit{Mdn})$^{b,d}$ & 35--44y & 35--44y & 25--34y & 25--34y & 5.091 & .165\\
\midrule
    & \multicolumn{6}{l}{After quality assurance measures (\textit{N} = 158)}\\
\cline{2-7}
    & \textit{control} & \textit{upward} & \textit{downward} & \textit{mixed} & \textit{H} value$^a$ & \textit{p} value\\ 
\hline
\textit{N}   &  40 & 39 & 38 & 41 & .. & ..\\
Gender$^{b,c}$ & 20f/20m & 20f/19m & 19f/19m & 20f/19m/1nb & 0.214 & .975 \\
Age (\textit{Mdn})$^{b,d}$ & 35--44y & 35--44y & 25--34y & 25--34y & 7.388 & .061\\
\bottomrule
\multicolumn{7}{l}{$^a$ non-parametric Kruskal–Wallis \textit{H} test}\\
\multicolumn{7}{l}{$^b$ note that one participant did not disclose age or gender information}\\
\multicolumn{7}{l}{$^c$ f = female, m = male, nb = non-binary / gender non-conforming}\\
\multicolumn{7}{l}{\makecell[l]{$^d$ \textit{Mdn} = median age (options: 18-24y, 25-34y, 35-44y, 45-54y, 55-64y, 65y and over)}}
\end{tabularx}
\end{table}

\subsection{Experimental Procedure}\label{subsec:experimental-procedure}

A detailed account of procedure and design choices underlying the experimental framework is given in~\cite{kuhl2023let}.

\begin{figure}[ht!]
\centering
\includegraphics[width=0.85\textwidth]{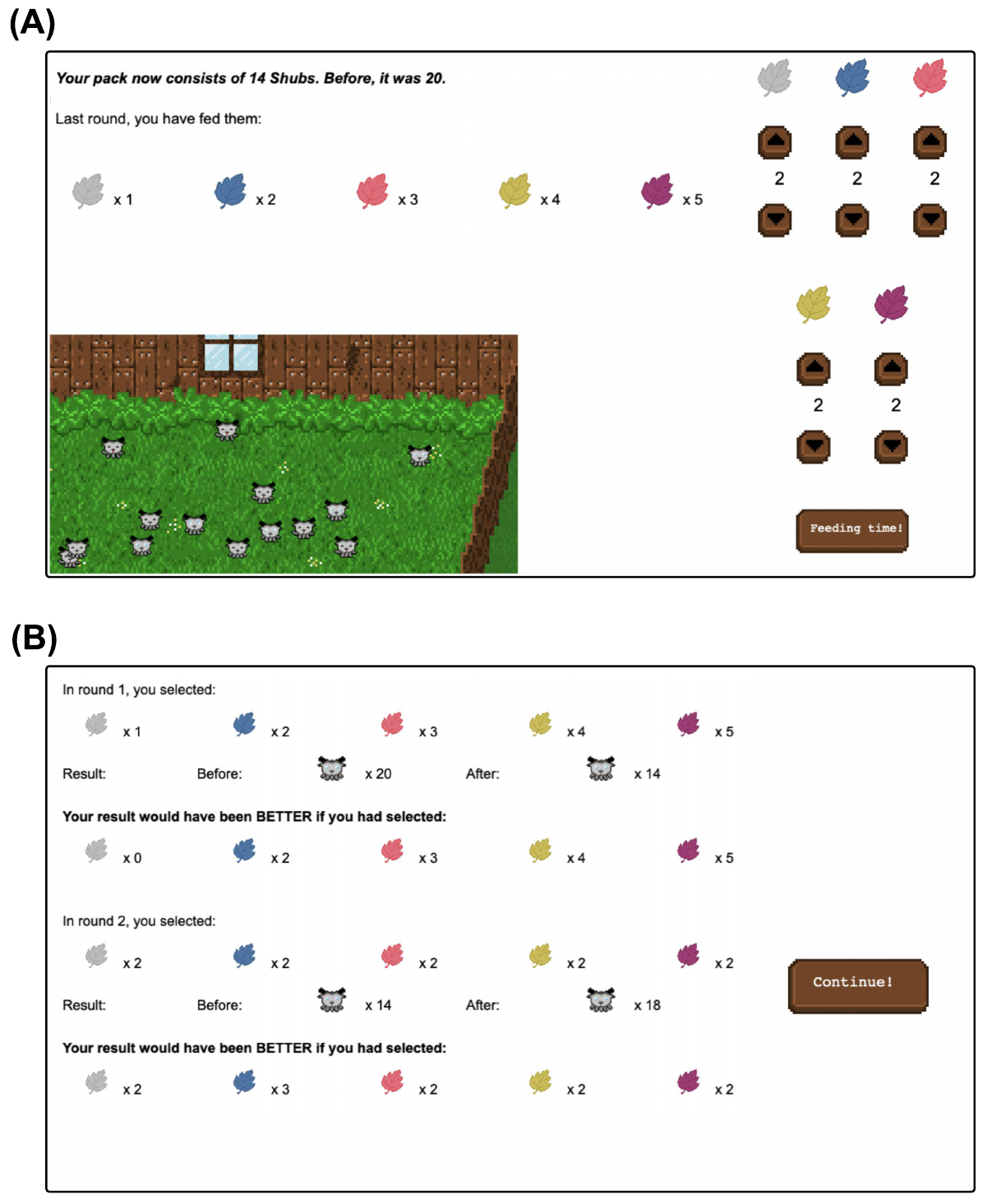}
\caption{Exemplary scenes from the Alien Zoo game. To improve visibility for this paper, font size in selected images was increased. \textbf{(A)} Example of a typical decision scene. Users are provided with a summary of their last choice, together with the previous and current pack size (note that the aliens are called `Shubs' in the experimental scenario). Moreover, the page shows a padlock with animated aliens to visualize the current pack size. 
The right side of the screen shows the plant types alongside upward and downward arrow buttons. Note that plant counters are set to 0 at the beginning of each new decision trial, the image above already shows the next selection (all plants set to 2).
\textbf{(B)} Example of a feedback scene for participants in the \textit{upward} \gls{CFE} condition, displaying user decision from the last two rounds, respective impact on alien number, and computed \gls{CFE}. Note that type of feedback varied depending on experimental group.} \label{fig:AZ}
\end{figure}

In short, users who agree to participate are directed to a web server to complete a short online game.
As part of the game, participants feed a group of aliens iteratively over several trials, choosing combinations of different plants as food for their aliens. During every feeding choice, users may select up to 6 instances of each of five plants represented by leaf symbols of identical shape but different color (see Figure \ref{fig:AZ}A for an exemplary decision scene).
Each player starts out with an initial pack size of 20 aliens.
Tasked to find the best plant combination that makes the alien pack grow instead of declining, 
top players generating the highest number of aliens per experimental group received an additional monetary bonus.

To facilitate the learning process of what plants make an effective alien diet, participants receive feedback in the form of \glspl{CFE} after every even trial. The feedback depends on the respective participant's condition (\textit{upward} = ``Your result would have been BETTER if you had selected:''; \textit{downward} = ``Your result would have been WORSE if you had selected:''; \textit{mixed} = ``Your result would have been BETTER / WORSE if you had selected:''; \textit{control} = no explanation beyond an overview of past choices). Figure \ref{fig:AZ}B depicts an exemplary feedback scene for a participant in the \textit{upward} group.
The game phase consists of 12 trials (\ie, 12 feeding choices), with two attention checks assessing user's attentiveness after trials 3 and 7. 

Following the game phase, a survey assessed users' subjective judgments of presented feedback (via a modified version of the \gls{SCS}~\cite{holzinger_measuring_2020}).
On top of two items assessing explicit knowledge of feature relevance for task success, this scale measures the extent to which an \gls{XAI} system provides clear and understandable explanations for its decisions.
Users rate the quality of explanations based on factors such as completeness, consistency, relevance, and comprehensibility on a five-point Likert scale.
Finally, we collected demographic information on participants' age and gender, before users received a link to access full debriefing information.

\subsection{Prediction of New Alien Number and Generation of CFE Feedback}\label{subsubsec:models}

During the game, an underlying \gls{ML} model trained on simulated plant-growth-rate data determines changes in alien number. 
In each trial, the participant's feeding choice is passed on to a decision tree regression model~\cite{shalev-shwartz_understanding_2014} to predict a change rate for the current pack size (-10 to +10 aliens per decision, capped not to go below 2).
Here, we use the model and training data from a previous study relying on the same experimental framework (maximal tree depth of 5 with Gini splitting rule of CART~\cite{breiman_classification_1984}; see Experiment 2 from~\cite{kuhl2023let}).
This prior work demonstrated the feasibility of the experimental framework when comparing \textit{upward} \gls{CFE} feedback to a no-explanation \textit{control} condition, promising to yield similarly meaningful insights into potential effects driven by different types of \glspl{CFE}.
In addition, the choice to rely on freely-available material that was previously published provided us with exemplary data distributions to obtain realistic estimates for fixed and random effects for the \textit{a priori} power analysis.

The corresponding training data entails a dependency between two features (plant 2 and plant 4, respectively) and the output variable. 
Specifically, the growth rate scales linearly with values 1 to 5 for plant 2, iff plant 4 has a value of 1 or 2. 
To prevent users from applying a simple `the more, the better' approach, the dependence between growth rate and the value 6 of plant 2 was disrupted.

Together with each prediction, we also compute a \gls{CFE} presenting an alternative plant combination that differs minimally from the current input via optimization~\cite{wachter2017counterfactual}. 
In our implementation, a \gls{CFE} $\xcf\in\RN^\dimsym$ of an \gls{ML} model $\classifier:\RN^\dimsym\to\setY$ is realized as solving:

\begin{equation}\label{eq:counterfactual_opt_original}
\underset{\xcf \,\in\, \RN^\dimsym}{\arg\min}\; \loss\big(\classifier(\xcf), \ycf\big) + C \cdot \regularization(\xcf, \x)
\end{equation}

where $\x\in\RN^\dimsym$ denotes the original input, the regularization $\regularization(\cdot)$ penalizes deviations from the original input $\x$ (weighted by a regularization strength $C>0$), $\ycf\in\setY$ denotes the requested output/behavior of the model $\classifier(\cdot)$ under the counterfactual $\xcf$, and $\loss(\cdot)$ denotes a loss function penalizing deviations from the requested prediction. Thus, returned \glspl{CFE} correspond to minimal perturbations to the model's input that alter the final prediction to a desired outcome.
Given the regularization term $\regularization(\cdot)$, generated \glspl{CFE} based on this definition remain as similar to the original input $\x$ as possible.

Depending on the participant's condition, computed \glspl{CFE} either increase (\textit{upward} condition, and odd trials of the \textit{mixed} condition) or decrease (\textit{downward} condition, and even trials of the \textit{mixed} condition) the current growth rate prediction by a few decimal points.
After two trials, participants receive these \glspl{CFE} as feedback to further improve their performance in the game (see Figure \ref{fig:AZ}B).

\subsection{Statistical Analysis}
We use R-4.1.1~\cite{r_core_team_r_2021} for all statistical analyses, with experimental condition (\textit{control}, \textit{upward}, \textit{downward}, \textit{mixed}) as independent variable.
Given our longitudinal design, employ linear mixed models for data analysis to effectively address the correlations that arise from multiple measurements taken from each participant~\cite{detry_analyzing_2016,muth_alternative_2016}.
We investigate systematic differences between experimental groups over the 12 feeding trials (R package: lme4 v.4\textunderscore 1.1-27.1)~\cite{bates_fitting_2015}, with alien pack size over trials as dependent variable, fixed effects of group, trial number and their interaction, and a by-subjects random intercept.
We compared model fits using the analysis of variance function (stats package, base R).
Effect sizes are reported as $\eta_{\text{p}}^{2}$ (R package: effectsize v.0.5)~\cite{ben-shachar_effectsize_2020}.
Pairwise estimated marginal means analysis followed-up significant main effects or interactions, Bonferroni corrected to account for multiple comparisons.
We report respective effect sizes in terms of Cohen's \textit{d}.

We analyze survey data based on item type.
Missing values (i.e., users responding ``I do not know.'' for items assessing explicit knowledge, or ``I prefer not to answer.'' for items assessing subjective experience) were removed prior to the survey analysis.

The first two items of the survey evaluate the user's explicit knowledge of feature relevance for successful task completion.
Our goal is to determine a comprehensive measure of user knowledge through rewards and penalties for correct and incorrect responses, respectively.
To achieve this, we calculate the number of plants correctly identified per participant (\ie, the number of matches between ground truth and user input).

The remaining items were adapted from the \gls{SCS}, a rating scale that allows users to evaluate the extent to which an \gls{XAI} system's explanations are clear, transparent, and understandable~\cite{holzinger_measuring_2020}.
Based on their responses to these items, we compute an adapted \gls{SCS} score for each participant to assess their subjective experience with the game.

Statistically, we investigate potential group differences concerning matches between user input and ground truth, Likert-style survey responses, age, and gender information using the non-parametric Kruskal–Wallis \textit{H} test (R package: rstatix v.0.7.0)~\cite{kassambara_rstatix_2021}, with effect sizes given as $\eta^{2}$.

Significant effects revealed via the Kruskal–Wallis \textit{H} test are followed-up by running pairwise comparisons between group levels, Bonferroni corrected for multiple testing.

\section{Results}

Overall, the results show group effects both in terms of performance during the game, and user's explicit knowledge of relevant and irrelevant features. 
However, we do not detect statistically significant differences when evaluating participants' subjective experience.

\subsection{Game Performance}

We evaluate users' game data to investigate whether naive users benefit comparably from different types of \glspl{CFE} when tasked to extract knowledge in an unfamiliar domain.
Specifically, we compare the number of aliens produced over time for participants receiving either \textit{upward} \glspl{CFE}, \textit{downward} \glspl{CFE}, \textit{mixed \glspl{CFE}}, or no \glspl{CFE} (\textit{control}) in the Alien Zoo iterative learning task.

\begin{figure}
\includegraphics[width=\textwidth]{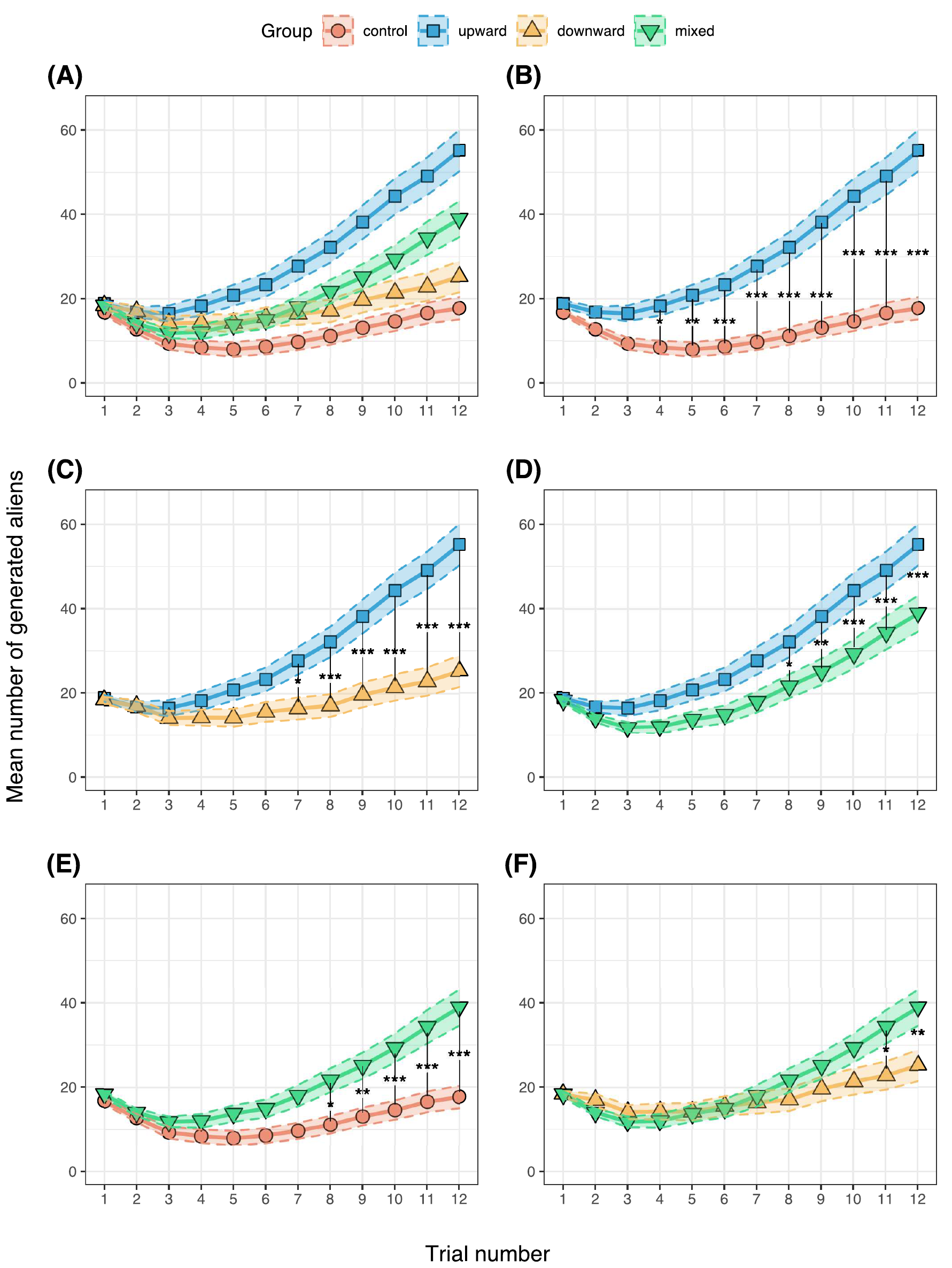}
\caption{Development of mean number of generated aliens per trial, plotted together for all groups \textbf{(A)}, and pairwise for those groups showing significant differences in the analysis following-up the significant interaction \textbf{(B-F)}.
Shaded areas denote the standard error of the mean.
Asterisks denote statistical significance with \textit{p} \textless .05 (*), \textit{p} \textless .01 (**), and \textit{p} \textless .001 (***), respectively.} \label{fig:performance}
\end{figure}

Figure \ref{fig:performance}A depicts the development of average pack size over trials.
All participants depict a positive learning trajectory, but with strikingly different slopes for different groups. 
The performance curves suggest that the mean number of generated aliens over trials varies as a function of experimental condition, with users receiving no explanations showing the least, and users receiving \textit{upward} \gls{CFE} feedback exhibiting the strongest performance increase.
The corresponding linear mixed effects model revealed a significant interaction between trial number and group (F(33,1694) = 13.114, \textit{p} \textless .001, $\eta_{\text{p}}^{2}$ = 0.203), confirming this observation. 

Follow-up analyses reveal an intriguing pattern of distinctive group differences (see Figures \ref{fig:performance}B-F).

The trajectories of the \textit{control} and the \textit{upward} group diverge significantly from trial 4 onward (t(300) $\geq$ -2.660, \textit{p} $\leq$ .0494, \textit{d} $\geq$ -1.066), with participants in the \textit{upward} group clearly outperforming \textit{control} participants (Figure \ref{fig:performance}B).
This pattern also holds when comparing the \textit{upward} group with the two remaining conditions. 
Trajectories of the \textit{upward} and the \textit{downward} group diverge significantly from trial 7 onward (t(300) $\geq$ 3.016, \textit{p} $\leq$ .0167, \textit{d} $\geq$ 1.224; Figure \ref{fig:performance}C).
Statistical differences between the \textit{upward} and the \textit{mixed} groups emerge starting at trial 8 (t(300) $\geq$ 2.851, \textit{p} $\leq$ .0280, \textit{d} $\geq$ 1.135; Figure \ref{fig:performance}D).

While performing less efficient as participants in the \textit{upward} group, participants in the \textit{mixed} condition also achieve statistically higher scores in the last 5 trials compared to \textit{control} participants (t(300) $\geq$ -2.891, \textit{p} $\leq$ .0247, \textit{d} $\geq$ -1.144; Figure \ref{fig:performance}E), and in the last 2 trials compared to \textit{downward} participants (t(300) $\geq$ -3.121, \textit{p} $\leq$ .0119, \textit{d} $\geq$ -1.251; Figure \ref{fig:performance}F). Only trajectories of participants in the \textit{control} and \textit{downward} conditions do not show any statistically meaningful differences. 

This interaction is complemented by a significant main effect of trial number (\textit{F}(11,1694) = 95.573, \textit{p} \textless .001, $\eta_{\text{p}}^{2}$ = 0.380), and group (\textit{F}(3,154) = 11.423, \textit{p} \textless .001, $\eta_{\text{p}}^{2}$ = 0.180).

\subsection{Assessing User's Explicit Knowledge}

\begin{figure}[t]
\includegraphics[width=\textwidth]{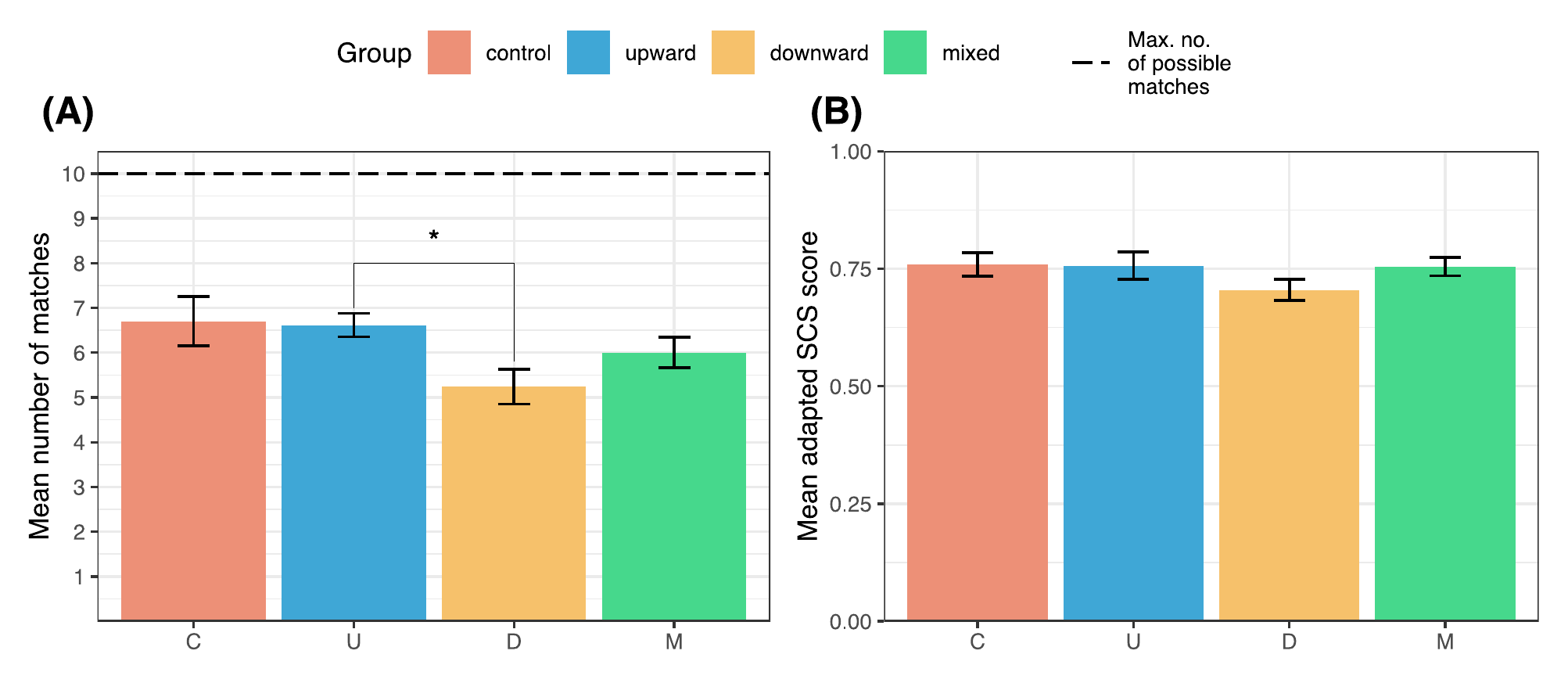}
\caption{\textbf{(A)} Mean number of matches between user input and ground truth in first two items of survey, assessing whether participants can correctly identify plants that are relevant and irrelevant for task success. The dashed line shows the maximally attainable number of matches (\ie, user responses perfectly aligned with ground truth). Error bars denote the standard error of the mean. The asterisk denotes statistical significance with \textit{p} \textless 0.05 (*). \textbf{(B)} Mean adapted \gls{SCS} scores across groups. Error bars denote the standard error of the mean.
} \label{fig:explicitKnow_SCS_combined}
\end{figure}

The first two items of the survey phase assess participants' explicit knowledge of feature relevance for task completion.
Across the two items, a participant could potentially reach 10 correct decisions by matching up their responses with the ground truth perfectly.
In terms of mean number of matches between ground truth and user judgments, participants in the \textit{control} condition matched highest (\textit{M} = 6.700 $\pm$ 0.548 \textit{SE}), followed by participants in the \textit{upward} (\textit{M} = 6.615 $\pm$ 0.261 \textit{SE}), \textit{mixed} (\textit{M} = 6.000 $\pm$ 0.342 \textit{SE}), and \textit{downward} condition (\textit{M} = 5.241 $\pm$ 0.390 \textit{SE}; Figure \ref{fig:explicitKnow_SCS_combined}A). The corresponding statistical analysis reveals a significant effect of group (\textit{H}(3) = 10.9, \textit{p} = .012, $\eta^{2}$ = 0.077).
Follow-up pairwise comparisons show that participants in the \textit{upward} match significantly higher than participants in the \textit{downward} condition (\textit{p} = .028).

\subsection{Assessing User's Subjective Experience}

A modified version of the \gls{SCS} informs whether participants perceive provided explanations as clear, understandable, and usable. 
As shown in Figure \ref{fig:explicitKnow_SCS_combined}B, participants in the \textit{control} (\textit{M} = 0.760 $\pm$ 0.025 \textit{SE}),
\textit{upward} (\textit{M} = 0.756 $\pm$ 0.029 \textit{SE}), and \textit{mixed} (\textit{M} = 0.754 $\pm$ 0.019 \textit{SE}) conditions achieve very similar scores, while the mean \gls{SCS} score of \textit{downward} participants is slightly lower (\textit{M} = 0.705 $\pm$ 0.023 \textit{SE}).
There is no statistically significant effect of group in terms of \gls{SCS} scores (\textit{H}(3) = 5.36, \textit{p} = .147, $\eta^{2}$ = 0.017).

\section{Discussion}

The current study investigates the impact of directionality of \glspl{CFE} for \gls{XAI} on objective task performance, explicit knowledge, and subjective experience of novice users during an iterative learning paradigm in an unfamiliar domain.

The results suggest that participants benefit most from receiving \textit{upward} \gls{CFE} feedback (\ie, informing them what choices would have been better), outperforming participants in all other conditions (Figure \ref{fig:performance}).
Consequently, we replicate prior work showing that \textit{upward} \glspl{CFE} induce a significant performance advantage over a no-explanation \textit{control}~\cite{kuhl2023let} in the employed experimental framework, and extend previous insights by the aspect of directionality.
In the current experimental setting, \textit{upward} counterfactuals may have provided novice users with interpretable and clear pathways for actions that improve future behavior~\cite{byrne_counterfactuals_2019}.
This is in line with previous psychological research demonstrating that reflecting upon ``better worlds'' may serve as a driving force for learning and adapting behavior~\cite{roese1994functional,wong2007narrating,epstude2008functional,myers2014role}.
Given the current task, the striking positive impact of \textit{upward} \glspl{CFE} is in line with the psychological concept of regulatory fit, as describing how a choice would have been better matches the motivational orientation to improve one's feeding choices~\cite{higgins2000making}.
Previous work in various domains suggests that such a feeling of fit induces more effective and satisfying performance, as well as greater persistence and motivation to continue the task~\cite{higgins2000making,motyka2014regulatory,ludolph2015does}. 
A similar mechanism may be in effect in the current setting.

Intriguingly, participants who receive \textit{mixed} \gls{CFE} feedback also achieve statistically higher scores compared to \textit{control} and \textit{downward} groups, specifically towards later trials. 
Considering that \textit{downward} \glspl{CFE} do not improve user performance compared to providing no explanations at all (\textit{control}), we may suspect that users in the \textit{mixed} condition benefit from receiving feedback that partially possesses regulatory fit. 

A previous study suggests a beneficial effect of explanations that establish a downward comparison~\cite{celar2023people}.
However, other research shows that \textit{downward} \glspl{CFE} are relatively more disliked compared to feature rankings in terms of user preference~\cite{ramon2021understanding}.
Intriguingly, participants receiving \textit{downward} \glspl{CFE} in the current work do not show statistically meaningful differences in terms of task performance compared to participants receiving no explanations at all. 
While the current discrepancy of the \textit{downward} group from the performance of the two other \gls{CFE} groups is striking, it merits only cautious interpretation generally devoted to null effects.
On the one hand, \textit{downward} \gls{CFE} feedback may have simply induced complacency that impeded participants' motivation to act, knowing that there still was a worse route they could have taken~\cite{roese1994functional,mcmullen2000downward}.
On the other hand, the current scenario may have been inadequate to unleash the negative affect necessary to stimulate action through the presented \textit{downward} comparisons. 
Unsuccessfully feeding aliens had only limited personal consequences for participants, potentially keeping the level of perceived regret following a sub-optimal decision comparatively low.
Consequently, the beneficial impact of \textit{downward} \glspl{CFE} in terms of regret minimization could not be observed~\cite{parikh2022efficacy}.
A future study could investigate this possibility more closely via an adapted design that involves increased personal costs, thereby implementing a penalty for poor decision-making and a higher chance of inducing regret.

In terms of explicit knowledge, participants in the \textit{upward} condition identified relevant and irrelevant input features more readily than in the \textit{downward} condition, in line with the performance advantage for \textit{upward} \glspl{CFE}.
This suggests that -- in tasks requiring users to extract new information from a system -- \textit{upward} \glspl{CFE} may be the better option for enhancing user's explicit knowledge.
A curious detail meriting comment, however, concerns the comparably high performance in terms of explicit knowledge by \textit{control} users that do not receive explanations at all. 
This may be explained by the relatively high proportion of \textit{control} participants indicating that they ``do not know'' for items assessing explicit knowledge, thus being discarded from this analysis. 
The remaining data may represent \textit{control} individuals who are the most knowledgeable and confident in their responses.

In terms of objectively quantifiable measures, our study found tangible behavioral group differences, in stark contrast to user responses concerning the subjective usability of explanations provided.
This observation is consistent with the literature on the mismatch between these two measures~\cite{kuhl2022keep,warren2023categorical}, and further highlights the need to carefully consider both subjective and objective measures when evaluating \gls{XAI} approaches.

\subsection{Limitations}

In order to provide a comprehensive evaluation of the current findings, it is important to acknowledge and address limitations inherent in this study.

Given the experimental Alien Zoo framework, the results are obtained in relation to a very specific context and with a specific task, diverging from many real-life domains.
Today, we are already witnessing the significant impact of AI-based decision-making systems across a wide range of domains, including but not limited to health care~\cite{rajpurkar2022ai}, the legal system~\cite{mowbray2019utilizing}, and human resource management~\cite{votto2021artificial}.
We carefully considered the trade-offs involved in selecting a specific context.
Ultimately, the primary objective of investigating the usability and impact of counterfactual directionality on user behavior and experience, motivated the choice for a single and quite abstract domain (\ie, feeding aliens).
This set-up allowed us to maintain a high level of control over experimental variables to isolate effects driven by directionality of \glspl{CFE}, while minimizing confounding variables that could arise from varying contexts. 
Importantly, participants could engage with the counterfactual explanations and extract new knowledge from the automated system without being influenced by pre-existing domain knowledge.
Thus, the current approach facilitated a more detailed analysis of how \gls{CFE} directionality specifically affects the task at hand and the extraction of new knowledge from an automated system.
Uncovering these specific dynamics within a well-defined context provides a first step, laying out the foundation for future work.
Results await validation across various domains, tasks, and user populations, to contribute to a more comprehensive understanding of the broader applicability and usefulness of \glspl{CFE} across different scenarios.

Similarly, the current work does not cover different approaches for generating the \glspl{CFE}. 
As outlined in~\refeq{eq:counterfactual_opt_original}, we follow an optimization approach to generate minimal adjustments to the model’s input that -- depending on experimental condition -- either increase or decrease the predicted growth rate.
This approach is based on the initial definition of \gls{CFE} generation for \gls{ML}~\cite{wachter2017counterfactual} and various methods expand on the idea of using optimization principles for generating \glspl{CFE}~\cite{mc2018interpretable,sharma2019certifai,mothilal2020explaining}.
Alternative approaches generate counterfactual instances based on,~\eg, reinforcement learning~\cite{samoilescu2021model,verma2022amortized} or conditional generative adversarial networks~\cite{van2021conditional,yang2021model}. 
It is quite conceivable that the respective method for generating counterfactual explanations could indeed influence the final results. Different methods may introduce variations in the characteristics, interpretability, and quality of the explanations. Therefore, we have taken great care to select an optimization-based approach that aligns with established practices in the field.

A further potential confound of the design may be that it favors early discovery of an effective strategy, resulting in better performance over the duration of the experiment as the performance measure (number of generated aliens) accumulates over time.
Finally, the current study neglects to account for individual user characteristics. It may be that anxiety-prone individuals respond more strongly to \textit{downward} \gls{CFE} feedback, given altered emotional and probabilistic appraisal of \textit{upward} counterfactual thinking in individuals with high levels of trait anxiety~\cite{parikh2020phenomenology}.
Thus, further research is necessary to obtain a more comprehensive understanding of the role of directionality of \glspl{CFE} in \gls{XAI}.

\subsection{Contribution to Knowledge for xAI}

The findings presented in this study have significant implications for the field of explainable and trustworthy artificial intelligence. \glspl{CFE} have emerged as a popular approach in \gls{XAI}, as they provide insights into the changes required in input data to influence a model's output.
This study specifically focuses on the directionality of \glspl{CFE}, distinguishing between \textit{upward} counterfactuals (describing scenarios better than the factual state) and \textit{downward} counterfactuals (describing scenarios worse than the factual state).

Our results demonstrate the importance of \gls{CFE} directionality in shaping behavior and experience of novice users when interacting with an unknown automated system in an unfamiliar domain to extract new knowledge.
The findings indicate that \textit{upward} \glspl{CFE} offer a significant performance advantage over other forms of counterfactual feedback in the given explanation context.
Specifically, users were able to extract new knowledge more effectively and demonstrated higher explicit knowledge of the system when provided with \textit{upward} \glspl{CFE} compared to \textit{downward} \glspl{CFE}.

These findings point towards critical role of regulatory fit in determining the effectiveness of model explanations~\cite{motyka2014regulatory}.
Regulatory fit refers to the alignment between an explanation and the task at hand. In the context of \gls{XAI}, this implies that the directionality of \glspl{CFE} should be carefully considered to ensure they are relevant and meaningful to the users' objectives and cognitive processes.
By providing explanations that align with users' goals and expectations, \gls{XAI} systems can enhance user performance and improve their understanding of the underlying models~\cite{miller_explanation_2019}.

The impact of these findings on \gls{XAI} as a sub-field of artificial intelligence is substantial.
\gls{XAI} aims to bridge the gap between black-box models and human comprehension, enabling users to trust, interpret, and interact with automated systems more effectively.
By identifying the advantages of \textit{upward} \glspl{CFE} and the potential benefits of \textit{mixed} \glspl{CFE}, this study contributes to the development of more effective and user-centric explainability techniques.
Understanding the directionality of \glspl{CFE} provides valuable insights into how explanations can be tailored to meet users' needs and improve their decision-making processes.

Furthermore, these findings have broader implications for the wider \gls{XAI} community. Researchers and practitioners in \gls{XAI} can leverage this knowledge to design better explainable systems.
They can incorporate the directionality of \glspl{CFE} into the design of \gls{XAI} interfaces, ensuring that explanations are presented in a way that maximizes user understanding and performance.
Additionally, these findings highlight the importance of user-centric evaluation methodologies in \gls{XAI} research, as they provide valuable insights into the impact of explanations on user behavior and knowledge acquisition.

\subsection{Conclusion}

The canonical example illustrating the concept of counterfactuals in \gls{XAI} is an \textit{upward} \gls{CFE}: ``If you had done X, your loan would have been approved.''
The current results, suggesting that \textit{upward} \glspl{CFE} are most effective for guiding decision-making, may explain why this example is considered to be an inherently intuitive prototype.
Further, the results of this study provide renewed evidence for the importance of considering not only algorithmic aspects of explainability approaches, but also their effectiveness during hands-on human-system interaction.
Specifically, they give reason to assume that regulatory fit, \ie, the alignment between an explanation and the task at hand, may act as a potentially crucial factor in determining the effectiveness of model explanations.

%
%
%
\newpage
\bibliographystyle{splncs04}
\bibliography{DAZ_bibliography} 

\begin{thebibliography}{10}
\providecommand{\url}[1]{\texttt{#1}}
\providecommand{\urlprefix}{URL }
\providecommand{\doi}[1]{https://doi.org/#1}

\bibitem{artelt2019computation}
Artelt, A., Hammer, B.: On the computation of counterfactual explanations--a
  survey. arXiv preprint arXiv:1911.07749  (2019)

\bibitem{artelt2021evaluating}
Artelt, A., Vaquet, V., Velioglu, R., Hinder, F., Brinkrolf, J., Schilling, M.,
  Hammer, B.: Evaluating robustness of counterfactual explanations. In: 2021
  IEEE Symposium Series on Computational Intelligence (SSCI). pp. 01--09. IEEE
  (2021). \doi{10.1109/SSCI50451.2021.9660058}

\bibitem{bansal_updates_2019}
Bansal, G., Nushi, B., Kamar, E., Weld, D.S., Lasecki, W.S., Horvitz, E.:
  Updates in {Human}-{AI} {Teams}: {Understanding} and {Addressing} the
  {Performance}/{Compatibility} {Tradeoff}. Proceedings of the AAAI Conference
  on Artificial Intelligence  \textbf{33},  2429--2437 (Jul 2019).
  \doi{10.1609/aaai.v33i01.33012429}

\bibitem{bates_fitting_2015}
Bates, D., M{\"a}chler, M., Bolker, B., Walker, S.: Fitting {Linear}
  {Mixed}-{Effects} {Models} {Using} \textbf{lme4}. Journal of Statistical
  Software  \textbf{67}(1) (2015). \doi{10.18637/jss.v067.i01}

\bibitem{ben-shachar_effectsize_2020}
Ben-Shachar, M., L{\"u}decke, D., Makowski, D.: effectsize: {Estimation} of
  {Effect} {Size} {Indices} and {Standardized} {Parameters}. Journal of Open
  Source Software  \textbf{5}(56), ~2815 (Dec 2020). \doi{10.21105/joss.02815}

\bibitem{breiman_classification_1984}
Breiman, L., Friedman, J.H., Olshen, R.A., Stone, C.J.: Classification {And}
  {Regression} {Trees}. Routledge, London, UK, 1 edn. (1984).
  \doi{10.1201/9781315139470}

\bibitem{byrne_counterfactuals_2019}
Byrne, R.M.: Counterfactuals in explainable artificial intelligence (xai):
  Evidence from human reasoning. In: Proceedings of the Twenty-Eighth
  International Joint Conference on Artificial Intelligence, {IJCAI-19}. pp.
  6276--6282. International Joint Conferences on Artificial Intelligence
  Organization (7 2019). \doi{10.24963/ijcai.2019/876}

\bibitem{byrne2016counterfactual}
Byrne, R.M.: Counterfactual thought. Annual review of psychology  \textbf{67},
  135--157 (2016). \doi{10.1146/annurev-psych-122414-033249}

\bibitem{celar2023people}
Celar, L., Byrne, R.M.: How people reason with counterfactual and causal
  explanations for artificial intelligence decisions in familiar and unfamiliar
  domains. Memory \& Cognition pp. 1--16 (2023).
  \doi{10.3758/s13421-023-01407-5}

\bibitem{dai2022counterfactual}
Dai, X., Keane, M.T., Shalloo, L., Ruelle, E., Byrne, R.M.: Counterfactual
  explanations for prediction and diagnosis in xai. In: Proceedings of the 2022
  AAAI/ACM Conference on AI, Ethics, and Society. pp. 215--226 (2022).
  \doi{10.1145/3514094.3534144}

\bibitem{detry_analyzing_2016}
Detry, M.A., Ma, Y.: Analyzing {Repeated} {Measurements} {Using} {Mixed}
  {Models}. JAMA  \textbf{315}(4), ~407 (Jan 2016).
  \doi{10.1001/jama.2015.19394}

\bibitem{dyczewski2012general}
Dyczewski, E.A., Markman, K.D.: General attainability beliefs moderate the
  motivational effects of counterfactual thinking. Journal of Experimental
  Social Psychology  \textbf{48}(5),  1217--1220 (2012).
  \doi{10.1016/j.jesp.2012.04.016}

\bibitem{epstude2008functional}
Epstude, K., Roese, N.J.: The functional theory of counterfactual thinking.
  Personality and social psychology review  \textbf{12}(2),  168--192 (2008)

\bibitem{goldinger_blaming_2003}
Goldinger, S.D., Kleider, H.M., Azuma, T., Beike, D.R.: "{Blaming} {The}
  {Victim}" {Under} {Memory} {Load}. Psychological Science  \textbf{14}(1),
  81--85 (Jan 2003). \doi{10.1111/1467-9280.01423}

\bibitem{higgins2000making}
Higgins, E.T.: Making a good decision: value from fit. American psychologist
  \textbf{55}(11), ~1217 (2000). \doi{10.1037/0003-066X.55.11.1217}

\bibitem{hilton_knowledge-based_1986}
Hilton, D.J., Slugoski, B.R.: Knowledge-based causal attribution: {The}
  abnormal conditions focus model. Psychological Review  \textbf{93}(1),
  75--88 (1986). \doi{10.1037/0033-295X.93.1.75}

\bibitem{holzinger_measuring_2020}
Holzinger, A., Carrington, A., M{\"u}ller, H.: Measuring the {Quality} of
  {Explanations}: {The} {System} {Causability} {Scale} ({SCS}): {Comparing}
  {Human} and {Machine} {Explanations}. KI - K{\"u}nstliche Intelligenz
  \textbf{34}(2),  193--198 (Jan 2020). \doi{10.1007/s13218-020-00636-z}

\bibitem{karimi2020model}
Karimi, A.H., Barthe, G., Balle, B., Valera, I.: Model-agnostic counterfactual
  explanations for consequential decisions. In: International Conference on
  Artificial Intelligence and Statistics. pp. 895--905. PMLR (2020)

\bibitem{kassambara_rstatix_2021}
Kassambara, A.: rstatix: Pipe-Friendly Framework for Basic Statistical Tests
  (2021), \url{https://CRAN.R-project.org/package=rstatix}, r package version
  0.7.0

\bibitem{keane2020good}
Keane, M.T., Smyth, B.: Good counterfactuals and where to find them: A
  case-based technique for generating counterfactuals for explainable ai (xai).
  In: Case-Based Reasoning Research and Development: 28th International
  Conference, ICCBR 2020, Salamanca, Spain, June 8--12, 2020, Proceedings 28.
  pp. 163--178. Springer (2020).
  \doi{10.1007/978-3-030-58342-2\textunderscore11}

\bibitem{kuhl2022keep}
Kuhl, U., Artelt, A., Hammer, B.: Keep your friends close and your
  counterfactuals closer: Improved learning from closest rather than plausible
  counterfactual explanations in an abstract setting. In: 2022 ACM Conference
  on Fairness, Accountability, and Transparency. pp. 2125--2137 (2022).
  \doi{10.1145/3531146.3534630}

\bibitem{kuhl2023let}
Kuhl, U., Artelt, A., Hammer, B.: Let's go to the alien zoo: Introducing an
  experimental framework to study usability of counterfactual explanations for
  machine learning. Frontiers in Computer Science  \textbf{5}, ~20 (2023).
  \doi{10.3389/fcomp.2023.1087929}

\bibitem{kumle2021estimating}
Kumle, L., V{\~o}, M.L.H., Draschkow, D.: Estimating power in (generalized)
  linear mixed models: An open introduction and tutorial in r. Behavior
  research methods  \textbf{53}(6),  2528--2543 (2021).
  \doi{10.3758/s13428-021-01546-0}

\bibitem{lage2019human}
Lage, I., Chen, E., He, J., Narayanan, M., Kim, B., Gershman, S.J.,
  Doshi-Velez, F.: Human evaluation of models built for interpretability. In:
  Proceedings of the AAAI Conference on Human Computation and Crowdsourcing.
  vol.~7, pp. 59--67 (2019). \doi{10.1609/hcomp.v7i1.5280}

\bibitem{lombrozo_explanation_2012}
Lombrozo, T.: Explanation and {Abductive} {Inference}. In: Holyoak, K.J.,
  Morrison, R.G. (eds.) The {Oxford} {Handbook} of {Thinking} and {Reasoning},
  pp. 260--276. Oxford University Press, Oxford, UK (Mar 2012).
  \doi{10.1093/oxfordhb/9780199734689.013.0014}

\bibitem{ludolph2015does}
Ludolph, R., Schulz, P.J.: Does regulatory fit lead to more effective health
  communication? a systematic review. Social Science \& Medicine  \textbf{128},
   142--150 (2015). \doi{https://doi.org/10.1016/j.socscimed.2015.01.021}

\bibitem{lundberg2017unified}
Lundberg, S.M., Lee, S.I.: A unified approach to interpreting model
  predictions. Advances in neural information processing systems  \textbf{30}
  (2017)

\bibitem{markman1993mental}
Markman, K.D., Gavanski, I., Sherman, S.J., McMullen, M.N.: The mental
  simulation of better and worse possible worlds. Journal of experimental
  social psychology  \textbf{29}(1),  87--109 (1993).
  \doi{10.1006/jesp.1993.1005}

\bibitem{mc2018interpretable}
Mc~Grath, R., Costabello, L., Le~Van, C., Sweeney, P., Kamiab, F., Shen, Z.,
  Lecue, F.: Interpretable credit application predictions with counterfactual
  explanations. In: NIPS 2018-Workshop on Challenges and Opportunities for AI
  in Financial Services: the Impact of Fairness, Explainability, Accuracy, and
  Privacy (2018)

\bibitem{mcmullen2000downward}
McMullen, M.N., Markman, K.D.: Downward counterfactuals and motivation: The
  wake-up call and the pangloss effect. Personality and Social Psychology
  Bulletin  \textbf{26}(5),  575--584 (2000). \doi{10.1177/0146167200267005}

\bibitem{miller_explanation_2019}
Miller, T.: Explanation in artificial intelligence: {Insights} from the social
  sciences. Artificial Intelligence  \textbf{267},  1--38 (Feb 2019).
  \doi{10.1016/j.artint.2018.07.007}

\bibitem{mothilal2020explaining}
Mothilal, R.K., Sharma, A., Tan, C.: Explaining machine learning classifiers
  through diverse counterfactual explanations. In: Proceedings of the 2020
  conference on fairness, accountability, and transparency. pp. 607--617
  (2020). \doi{10.1145/3351095.3372850}

\bibitem{motyka2014regulatory}
Motyka, S., Grewal, D., Puccinelli, N.M., Roggeveen, A.L., Avnet, T., Daryanto,
  A., de~Ruyter, K., Wetzels, M.: Regulatory fit: A meta-analytic synthesis.
  Journal of Consumer Psychology  \textbf{24}(3),  394--410 (2014).
  \doi{10.1016/j.jcps.2013.11.004}

\bibitem{mowbray2019utilizing}
Mowbray, A., Chung, P., Greenleaf, G.: Utilizing ai in the legal assistance
  sector. In: LegalAIIA@ ICAIL. pp. 12--18 (2019)

\bibitem{muth_alternative_2016}
Muth, C., Bales, K.L., Hinde, K., Maninger, N., Mendoza, S.P., Ferrer, E.:
  Alternative {Models} for {Small} {Samples} in {Psychological} {Research}:
  {Applying} {Linear} {Mixed} {Effects} {Models} and {Generalized} {Estimating}
  {Equations} to {Repeated} {Measures} {Data}. Educational and Psychological
  Measurement  \textbf{76}(1),  64--87 (Feb 2016).
  \doi{10.1177/0013164415580432}

\bibitem{myers2014role}
Myers, A.L., McCrea, S.M., Tyser, M.P.: The role of thought-content and mood in
  the preparative benefits of upward counterfactual thinking. Motivation and
  Emotion  \textbf{38},  166--182 (2014). \doi{10.1007/s11031-013-9362-5}

\bibitem{nielsen1994measuring}
Nielsen, J., Levy, J.: Measuring usability: preference vs. performance.
  Communications of the ACM  \textbf{37}(4),  66--75 (1994).
  \doi{10.1145/175276.175282}

\bibitem{parikh2022efficacy}
Parikh, N., De~Brigard, F., LaBar, K.S.: The efficacy of downward
  counterfactual thinking for regulating emotional memories in anxious
  individuals. Frontiers in Psychology  \textbf{12},  712066 (2022).
  \doi{10.3389/fpsyg.2021.712066}

\bibitem{parikh2020phenomenology}
Parikh, N., LaBar, K.S., De~Brigard, F.: Phenomenology of counterfactual
  thinking is dampened in anxious individuals. Cognition and Emotion
  \textbf{34}(8),  1737--1745 (2020). \doi{10.1080/02699931.2020.1802230}

\bibitem{qiao2021learning}
Qiao, L., Wang, W., Lin, B.: Learning accurate and interpretable decision rule
  sets from neural networks. In: Proceedings of the AAAI Conference on
  Artificial Intelligence. vol.~35, pp. 4303--4311 (2021).
  \doi{10.1609/aaai.v35i5.16555}

\bibitem{r_core_team_r_2021}
{R Core Team}: R: {A} {Language} and {Environment} for {Statistical}
  {Computing}. R Foundation for Statistical Computing, Vienna, Austria (2021),
  \url{https://www.R-project.org/}

\bibitem{rajpurkar2022ai}
Rajpurkar, P., Chen, E., Banerjee, O., Topol, E.J.: Ai in health and medicine.
  Nature medicine  \textbf{28}(1),  31--38 (2022).
  \doi{10.1038/s41591-021-01614-0}

\bibitem{ramon2021understanding}
Ramon, Y., Vermeire, T., Toubia, O., Martens, D., Evgeniou, T.: Understanding
  consumer preferences for explanations generated by xai algorithms. arXiv
  preprint arXiv:2107.02624  (2021)

\bibitem{roese1994functional}
Roese, N.J.: The functional basis of counterfactual thinking. Journal of
  personality and Social Psychology  \textbf{66}(5), ~805 (1994).
  \doi{10.1037/0022-3514.66.5.805}

\bibitem{roese_counterfactual_1997}
Roese, N.J.: Counterfactual thinking. Psychological Bulletin  \textbf{121}(1),
  133--148 (1997). \doi{10.1037/0033-2909.121.1.133}

\bibitem{roese1995functions}
Roese, N.J., Olson, J.M.: Functions of counterfactual thinking. In: What Might
  Have Been: The Social Psychology of Counterfactual Thinking, pp. 169--197.
  Erlbaum (1995)

\bibitem{rozemberczki2022shapley}
Rozemberczki, B., Watson, L., Bayer, P., Yang, H.T., Kiss, O., Nilsson, S.,
  Sarkar, R.: The shapley value in machine learning. In: The 31st International
  Joint Conference on Artificial Intelligence and the 25th European Conference
  on Artificial Intelligence (2022). \doi{10.24963/ijcai.2022/778}

\bibitem{samoilescu2021model}
Samoilescu, R.F., Van~Looveren, A., Klaise, J.: Model-agnostic and scalable
  counterfactual explanations via reinforcement learning. arXiv preprint
  arXiv:2106.02597  (2021)

\bibitem{sanna_antecedents_1996}
Sanna, L.J., Turley, K.J.: Antecedents to {Spontaneous} {Counterfactual}
  {Thinking}: {Effects} of {Expectancy} {Violation} and {Outcome} {Valence}.
  Personality and Social Psychology Bulletin  \textbf{22}(9),  906--919 (Sep
  1996). \doi{10.1177/0146167296229005}

\bibitem{shalev-shwartz_understanding_2014}
Shalev-Shwartz, S., Ben-David, S.: Understanding machine learning: from theory
  to algorithms. Cambridge University Press, New York, NY, USA (2014)

\bibitem{sharma2019certifai}
Sharma, S., Henderson, J., Ghosh, J.: Certifai: Counterfactual explanations for
  robustness, transparency, interpretability, and fairness of artificial
  intelligence models. arXiv preprint arXiv:1905.07857  (2019)

\bibitem{shin2022prototype}
Shin, Y.M., Kim, S.W., Yoon, E.B., Shin, W.Y.: Prototype-based explanations for
  graph neural networks. In: Proceedings of the AAAI Conference on Artificial
  Intelligence. vol.~36, pp. 13047--13048 (2022).
  \doi{10.1609/aaai.v36i11.21660}

\bibitem{stepin2021survey}
Stepin, I., Alonso, J.M., Catala, A., Pereira-Fari{\~n}a, M.: A survey of
  contrastive and counterfactual explanation generation methods for explainable
  artificial intelligence. IEEE Access  \textbf{9},  11974--12001 (2021).
  \doi{10.1109/ACCESS.2021.3051315}

\bibitem{ustun2019actionable}
Ustun, B., Spangher, A., Liu, Y.: Actionable recourse in linear classification.
  In: Proceedings of the conference on fairness, accountability, and
  transparency. pp. 10--19 (2019). \doi{10.1145/3287560.3287566}

\bibitem{van2021conditional}
Van~Looveren, A., Klaise, J., Vacanti, G., Cobb, O.: Conditional generative
  models for counterfactual explanations. arXiv preprint arXiv:2101.10123
  (2021)

\bibitem{verma2020counterfactual}
Verma, S., Boonsanong, V., Hoang, M., Hines, K.E., Dickerson, J.P., Shah, C.:
  Counterfactual explanations and algorithmic recourses for machine learning: A
  review. arXiv preprint arXiv:2010.10596  (2020)

\bibitem{verma2022amortized}
Verma, S., Hines, K., Dickerson, J.P.: Amortized generation of sequential
  algorithmic recourses for black-box models. In: Proceedings of the AAAI
  Conference on Artificial Intelligence. vol.~36, pp. 8512--8519 (2022).
  \doi{10.1609/aaai.v36i8.20828}

\bibitem{votto2021artificial}
Votto, A.M., Valecha, R., Najafirad, P., Rao, H.R.: Artificial intelligence in
  tactical human resource management: A systematic literature review.
  International Journal of Information Management Data Insights  \textbf{1}(2),
   100047 (2021). \doi{10.1016/j.jjimei.2021.100047}

\bibitem{van2021evaluating}
van~der Waa, J., Nieuwburg, E., Cremers, A., Neerincx, M.: Evaluating xai: A
  comparison of rule-based and example-based explanations. Artificial
  Intelligence  \textbf{291},  103404 (2021).
  \doi{10.1016/j.artint.2020.103404}

\bibitem{wachter2017counterfactual}
Wachter, S., Mittelstadt, B., Russell, C.: Counterfactual explanations without
  opening the black box: Automated decisions and the gdpr. Harv. JL \& Tech.
  \textbf{31}, ~841 (2017)

\bibitem{warren2023categorical}
Warren, G., Byrne, R.M., Keane, M.T.: Categorical and continuous features in
  counterfactual explanations of ai systems. In: Proceedings of the 28th
  International Conference on Intelligent User Interfaces. pp. 171--187 (2023)

\bibitem{warren2022features}
Warren, G., Keane, M.T., , Byrne, R.M.: Features of explainability: How users
  understand counterfactual and causal explanations for categorical and
  continuous features in xai. In: IJCAI-ECAI’22 Workshop: Cognitive Aspects
  of Knowledge Representation (2022), https://ceur-ws.org/Vol-3251/paper1.pdf

\bibitem{white2005looking}
White, K., Lehman, D.R.: Looking on the bright side: Downward counterfactual
  thinking in response to negative life events. Personality and Social
  Psychology Bulletin  \textbf{31}(10),  1413--1424 (2005).
  \doi{10.1177/0146167205276064}

\bibitem{wong2007narrating}
Wong, E.M.: Narrating near-histories: The effects of counterfactual
  communication on motivation and performance. Management \& Organizational
  History  \textbf{2}(4),  351--370 (2007). \doi{10.1177/1744935907086119}

\bibitem{yang2021model}
Yang, F., Alva, S.S., Chen, J., Hu, X.: Model-based counterfactual synthesizer
  for interpretation. In: Proceedings of the 27th ACM SIGKDD conference on
  knowledge discovery \& data mining. pp. 1964--1974 (2021).
  \doi{10.1145/3447548.3467333}

\end{thebibliography}
%




\end{document}